\documentclass{article}
\usepackage[accepted]{icml2009}
\usepackage{graphicx}
\usepackage{subfigure}
\usepackage{mlapa}
\usepackage{amsmath,amsthm,amssymb}

\usepackage[latin9]{inputenc}
\usepackage{graphicx}
\usepackage{epsfig}

\newtheorem{thm}{Theorem}[section]
\newtheorem{defn}{Definition}[section]

\def\beq#1\eeq{\begin{equation}#1\end{equation}}
\def\beqa#1\eeqa{\begin{align}#1\end{align}}
\def\bseq#1\eseq{\begin{subequations}#1\end{subequations}}

\newcommand{\cS}{{\cal S}}

\newcommand{\xx}{{\bf x}}
\renewcommand{\ss}{{\bf s}}
\newcommand{\yy}{{\bf y}}
\renewcommand{\aa}{{\bf a}}
\newcommand{\bb}{{\bf b}}
\newcommand{\dd}{{\bf d}}
\renewcommand{\ll}{{\bf l}}
\newcommand{\Figref}[1]{Fig.~\ref{#1}}
\newcommand{\Eqref}[1]{Eq.~\ref{#1}} % The '~' stops breaking and gives correct spacing

\begin{document}

\twocolumn[
\icmltitle{Learning Nonlinear Dynamic Models}

% The author names and address list should only appear in the accepted version.
 \icmlauthor{John Langford}{jl@yahoo-inc.com}
 \icmladdress{Yahoo! Research, New York, NY 10011}
 \icmlauthor{Ruslan Salakhutdinov}{rsalakhu@cs.toronto.edu}
 \icmladdress{Department of Computer Science, University of Toronto, Ontario M2N6T3}
 \icmlauthor{Tong Zhang}{tongz@rci.rutgers.edu}
 \icmladdress{Department of Statistics, Rutgers University, Piscataway, NJ 08854}

\vskip 0.3in
]

%\maketitle

\begin{abstract}
  We present a novel approach for learning nonlinear dynamic models,
  which leads to a new set of tools capable of solving problems that
  are otherwise difficult.  We provide theory showing this new
  approach is consistent for models with long range structure, and
  apply the approach to motion capture and 
  high-dimensional video data, yielding
  results superior to standard alternatives.
\end{abstract}

\section{Introduction}

\begin{figure}[t]
  \centering
  \includegraphics[width=2.7in]{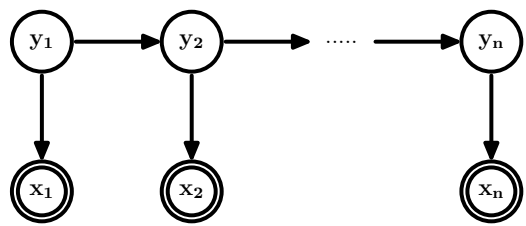}
  \caption{Dynamic Model with observation vector $\xx_t$ and hidden state 
   vector $\yy_t$.}
  \label{fig:dm}
\vspace{-0.2in} 
\end{figure}

The notion of hidden states appears in many nonstationary models of
the world such as Hidden Markov Models (HMMs), which have discrete
states, and Kalman filters, which have continuous states.
Figure~\ref{fig:dm} shows a general dynamic model with observation
$\xx_t$ and unobserved hidden state $\yy_t$.  The system is characterized
by a state transition probability $P(\yy_{t+1}|\yy_t)$, and a state to
observation probability $P(\xx_t|\yy_t)$.

The method for predicting future events under such a dynamic model is to
maintain a posterior distribution over the hidden state $\yy_{t+1}$, based on
all observations $X_{1:t}=\{\xx_1,\ldots,\xx_t\}$ up to time $t$. 
The posterior can be updated using the formula: 
\begin{align} 
&P(\yy_{t+1}|X_{1:t}) \nonumber\\
& \quad \propto \sum_{\yy_t} P(\yy_{t}|X_{1:{t-1}}) P(\xx_{t}|\yy_{t}) 
P(\yy_{t+1}|\yy_t)  .  \label{eq:post-update}
\end{align}
The prediction of future events $\xx_{t+1},\ldots,\xx_{t+k}$, $k>0$,
conditioned on $X_{1:t}$ is through the posterior over $\yy_t$:
\begin{align}
&P(\xx_{t+1},\ldots,\xx_{t+k}|\xx_{1:t}) \nonumber\\
& \quad \propto P(\yy_{t+1}|X_{1:t}) P(\xx_{t+1},\ldots,\xx_{t+k}|\yy_{t+1}) . \label{eq:predict}
\end{align}
Hidden state based dynamic models have a wide range of applications,
such as time series forecasting, finance, control,
robotics, video and speech processing.
Some detailed dynamic models and application examples
can be found in \cite{WestHarr97}.

From \Eqref{eq:predict}, it is clear that
the benefit of using a hidden state 
dynamic model is that the information contained
in the observation $X_{1:t}$ can be captured by a relatively
small hidden state $\yy_{t+1}$. 
Therefore in order to predict the future, we do not have
to use all previous observations $X_{1:t}$ but only its state representation
$\yy_{t+1}$. In principle, $\yy_{t+1}$ may contain a finite history of length
$k+1$, such as $\xx_{t},\xx_{t-1},\ldots,\xx_{t-k}$.
Although the notation only considers first order dependency, it incorporates
higher order dependency by considering a representation of the form
$Y_{t}=[\yy_t',\yy_{t-1}',\ldots,\yy_{t-k}']$, which is a standard trick.

In an HMM or Kalman filter, 
both transition and observation functions are linear maps. 
There are reasonable algorithms that can learn these linear dynamic models.
For example, in addition to the classical EM approach,
it was recently shown that global learning of certain
hidden Markov models can be achieved in polynomial time \cite{Linear}.
Moreover, for linear models, the posterior update rule is quite 
simple. Therefore, once the model parameters are estimated, such models
can be readily applied for prediction.

However in many real problems, the system dynamics cannot be
approximated linearly. For such problems, it is often necessary to 
incorporate nonlinearity into the dynamic model.
The standard approach to this problem is through nonlinear
probability modeling,
where prior knowledge is required to define a sensible state representation,
together with parametric forms of transition and observation probabilities.
The model parameters are learned by using probabilistic methods such as the EM
 \cite{WB,RG}.
When the learned model is applied for prediction purposes,
it is necessary to maintain the posterior
$P(\yy_{t}|X_{1:t})$ using the update
formula in \Eqref{eq:post-update}. Unfortunately, for nonlinear
systems, maintaining $P(\yy_{t}|\xx_{1:t})$ is generally difficult because
the posterior can become exponentially more complex (e.g., exponentially
many mixture components in a mixture model) as $t$ increases.

This computational difficulty is a significant obstacle to applying
nonlinear dynamic systems to practical problems. The traditional
approach to address the computational difficulty is through
approximation methods.  For example, in the particle filtering
approach \cite{GSS93,Arulampalam2002}, one uses a finite
number of samples to represent the posterior distribution and the
samples are then updated as observations arrive.  Another
%, arguably
%a more general 
approach is to maintain a mixture of Gaussians to
approximate the posterior, $P(\yy_{t}|X_{1:t})$, which may also be
regarded as a mixture of Kalman filters \cite{ChenLiu00}.  Although an
exponential in $t$ number of mixture components are needed to
accurately represent the posterior, in practice, one has to use a
fixed number of mixture components to approximate the distribution.
This leads to the following question: even if the posterior can be
well-approximated by a computationally tractable approximation family
(such as finite mixtures of Gaussians), how can one design a good
approximate inference method that is guaranteed to find a good quality
approximation?  
%Because of the complex techniques required to design a
%reasonable approximation scheme, the use of nonlinear dynamic models
%has been significantly limited for practical problems.
The use of complex techniques required to design 
reasonable approximation schemes makes it non-trivial to 
apply nonlinear dynamic models for many practical problems. 

This paper introduces an alternative approach, where we start with a
different representation of a linear dynamic model which we call the
{\em sufficient posterior representation}. It is shown that one can recover
the underlying state representation by using prediction methods that
are not necessarily probabilistic. This allows us to model nonlinear
dynamic behaviors with many available nonlinear supervised learning
algorithms such as neural networks, boosting, and support vector
machines in a simple and unified fashion. Compared to the traditional
approach, it has several distinct advantages:
\begin{itemize}
\item It does not require us to design any explicit state
  representation and probability model using prior knowledge.
  Instead, the representation is implicitly embedded in the
  representational choice of the underlying supervised learning
  algorithm, which may be regarded as a black box with the power to
  learn an arbitrary representation.  The prior knowledge can be
  simply encoded as input features to the learning algorithms, which
  significantly simplifies the modeling aspect.
\item It does not require us to come up with any specific
  representation of the posterior and the corresponding approximate
  Bayesian inference schemes for posterior updates. Instead, this
  issue is addressed by incorporating the posterior update as part of
  the learning process. Again, the posterior representation is
  implicitly embedded in the representational choice of the underlying
  supervised learning algorithm. In this sense, our scheme learns the
  optimal representation for posterior approximation and the
  corresponding update rules within the representational power of the
  underlying supervised algorithm\footnote{Many modern 
  supervised learning algorithms are universal, in
  the sense that they can learn an arbitrary representation in the
  large sample limit.}. 
%The power is unlimited in theory,
%  because many modern supervised learning algorithms are universal, in
%  the sense that they can learn an arbitrary representation in the
%  large sample limit.
\item It is possible to obtain performance guarantees for our
  algorithm in terms of the learning performance of the underlying
  supervised algorithm.  The performance of the latter has been
  heavily investigated in the statistical and learning theory
  literature. Such results can thus be applied to obtain theoretical
  results on our methods for learning nonlinear dynamic models.
\end{itemize}

%Because of the above mentioned advantages, we believe our approach
%opens up an entirely new set of devices for nonlinear dynamic
%modeling. It removes several obstacles in the traditional approach
%that requires heavy human design, and allows well-established
%supervised learning algorithms to be used automatically for nonlinear
%dynamic models. This leads to methods that can be readily used by
%people unfamiliar with sophisticated Bayesian inference techniques.

\section{Sufficient Posterior Representation}

Instead of starting with a probability model, our approach directly
attacks the problem of predicting $\yy_{t+k}$ based on $X_{1:t}$.
Clearly the prediction depends only on the posterior distribution
$P(\yy_{t+1}|X_{1:t})$. Therefore we can solve the prediction problem
as long as we can estimate, and update this posterior distribution.

\begin{figure}[t]
  \centering
  \includegraphics[width=2.5in]{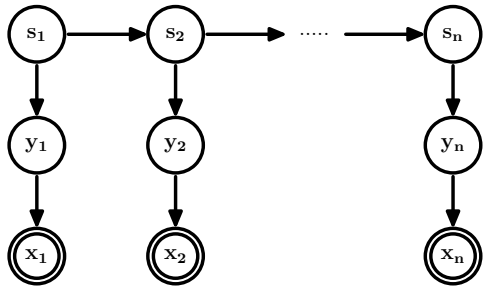} 
  \caption{Dynamic Model with observation vector $\xx_t$, 
hidden state vector $\yy_t$, and the posterior sufficient statistic 
 vector $\ss_t$.}
  \label{fig:pdm}
\vspace{-0.1in} 
\end{figure}

In our approach, it is assumed that the posterior 
$P(\yy_{t+1}|X_{1:t})$ 
can be approximated by a family of distributions parameterized
by $\ss_{t+1} \in \cS$: $P(\yy_{t+1}|X_{1:t}) \approx P(\yy_{t+1}|\ss_{t+1})$ for some
deterministic parameter $\ss_{t+1}$ that depends on $X_{1:t}$. That is, 
$\ss_{t+1}$ is a sufficient statistic for the posterior
$P(\yy_{t+1}|X_{1:t})$, and updating the posterior is equivalent to updating
the sufficient statistic~$\ss_{t+1}$.
The augmented model that incorporates the
(approximate) sufficient statistics $\ss_t \in \cS$ is shown in
\Figref{fig:pdm}.  In this model, $\yy_t$ can be integrated out, which
leaves a model containing only $\ss_t$ and $\xx_t$.

According to the posterior update of \Eqref{eq:post-update}, 
there exists a deterministic
function $B$ such that: 
\[
\ss_{{t+1}} = B(\xx_t,\ss_{t}) . 
\]
For simplicity, we can give an arbitrary value for the initial state $\ss_1$,
and let:
\[
\ss_2 = A(\xx_1) = B(\xx_1,\ss_1) .
\]
Moreover, according to \Eqref{eq:predict}, given an arbitrary
vector function $f$ of the future events 
$X_{t+1:\infty}=\{\xx_{t+1},\xx_{t+2},\cdots\}$,
there exists a deterministic
function $C^{f}$ ($k>0$) such that:
\[
E_{X_{t+1:\infty}} [f(X_{t+1:\infty}) | X_{1:t}] = C^{f} (\ss_{t+1}) .
\]
Therefore the dynamics of the model in \Figref{fig:dm}
is determined by the posterior initialization rule
$A$ and posterior update rule $B$. Moreover, the prediction of the
system is completely determined by the function $C^{f}$.

The key observation of our approach is that the functions
$A$, $B$, and $C$ are deterministic, which does not require any probability
assumption. It fully captures the correct 
dynamics of the underlying probabilistic dynamic model. 
However, by removing the probability assumption, we obtain a more general and 
flexible model. In particular, we are not required to start with specific
forms of the transition model $P(\yy_{t+1}|\yy_t)$, the observation model
$P(\xx_t|\yy_t)$, or the posterior sufficient statistic model
$P(\yy_{t+1}|\xx_{1:t}) \approx P(\yy_{t+1}|\ss_{t+1})$, as required in the standard approach.
Instead, we may embed the forms of such models into the functional 
approximation forms in standard learning algorithms, such as neural networks,
kernel machines, or tree ensembles. These are universal learning machines
that are well studied in the learning theory literature. 

Our approach essentially replaces a stochastic hidden state
representation through the actual state $Y$ by a deterministic
representation through the posterior sufficient statistic
$S$. Although the corresponding representation may become more
complex (which is why in the traditional approach, $\yy_t$ is always 
explicitly included in the model), this is not a problem in our approach, because we do not have
to know the explicit representation. Instead, the complexity is
incorporated into the underlying learning algorithm --- this allows us
to take advantage of sophisticated modern supervised learning
algorithms that can handle complex functional representations.
Moreover, unlike the traditional approach, in which one designs a
specific form of $P(\yy_t|\ss_t)$ by hand, and then derives an approximate
update rule $B$ by hand using Bayesian inference methods, here, we
simply use learning to come up with the best possible representation
and update (assuming the underlying learning algorithm is sufficiently
powerful). We believe this approach is also more robust because it is
less sensitive to model mis-specifications or non-optimal approximate
inference algorithms that commonly occur in practice.

By changing the standard probabilistic dynamic model in
\Figref{fig:dm} to its sufficient posterior representation in
\Figref{fig:pdm} (where we assume $\yy_t$ is integrated out, and
thus can be ignored), we can define the goal of our learning problem.
Since $\yy_t$ is removed from the formulation, in the following, we
shall refer to the sufficient posterior statistic $\ss_t$ simply as
state.

We can now introduce the following definition of Sufficient Posterior
Representation of Dynamic Model, which we refer to SPR-DM.
\begin{defn} (SPR-DM) \label{def:spr-dm}
A sufficient posterior representation of a dynamic model is
given by an observed sequence $\{\xx_t\}$ and unobserved hidden
state $\{\ss_t\}$, characterized by state initialization map
$\ss_2 = A(\xx_1)$, state update map $\ss_{t+1}=B(\xx_t,\ss_t)$, and
state prediction maps:
\[
E_{X_{t+1:\infty}} [f(X_{t+1:\infty}) | X_{1:t}] = C^{f} (\ss_{t+1}) 
\]
for any pre-determined vector function $C^{f}$.
\end{defn}

Our goal in this model is to learn the model dynamics characterized by
$A$ and $B$, as well as $C^f$ for any given vector function of
interest.

\begin{figure*}
\centering
\includegraphics[width=1.6in]{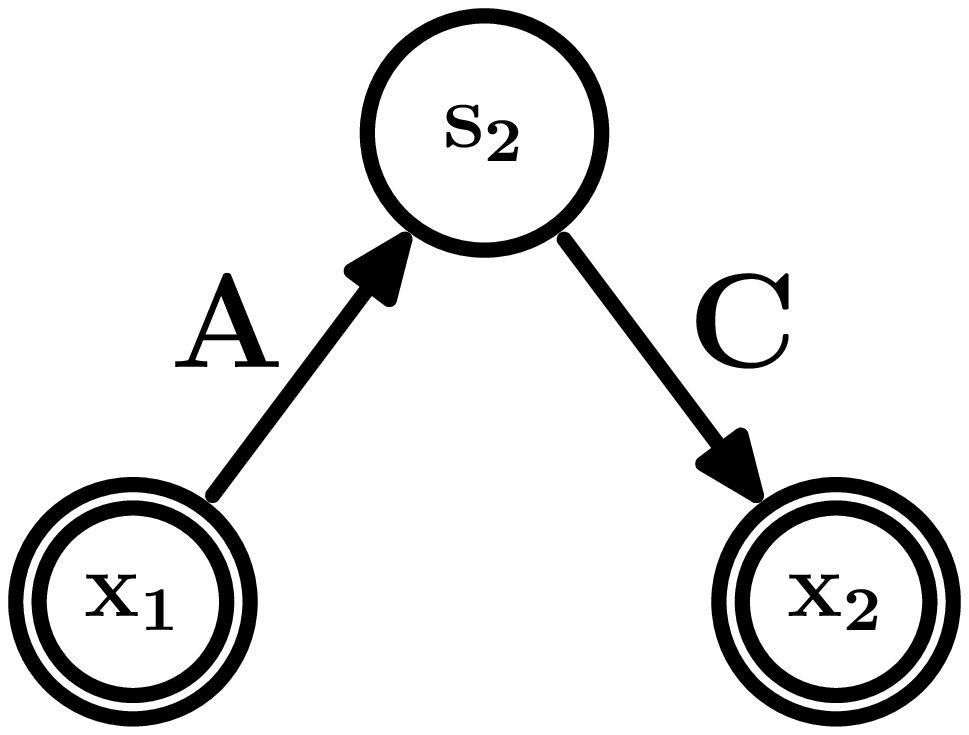} \hspace{0.1in}
\includegraphics[width=1.9in]{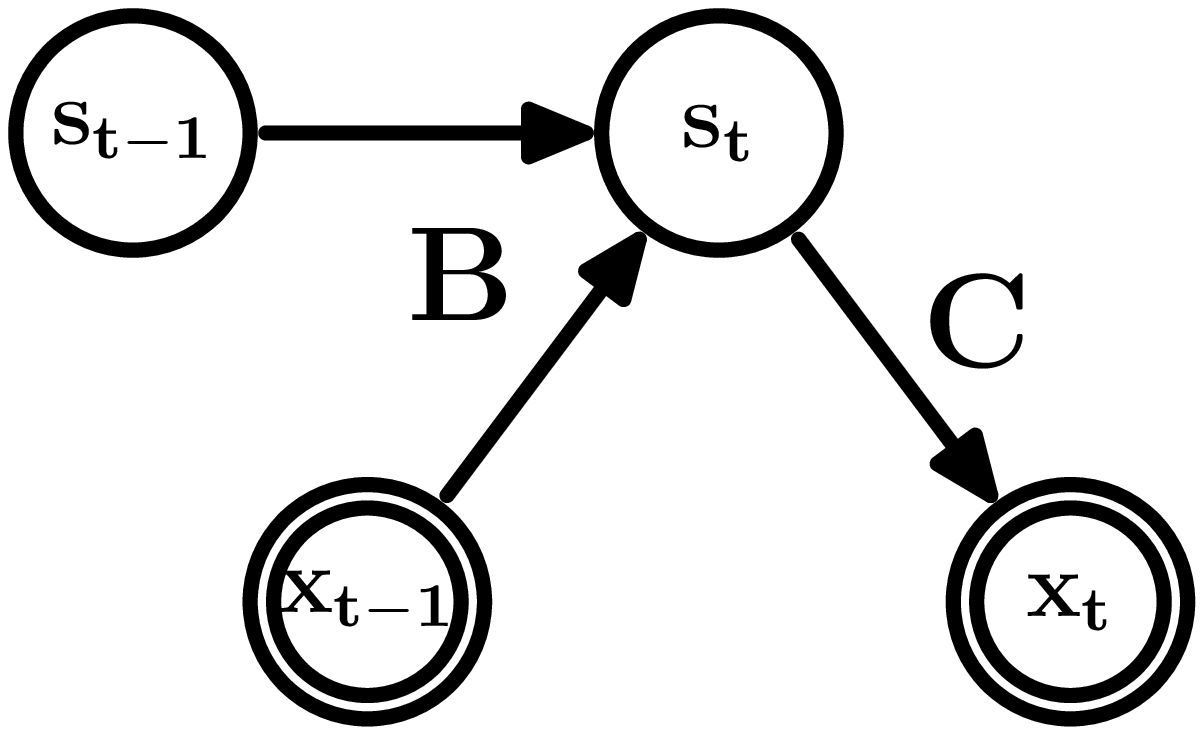} \hspace{0.1in}
\includegraphics[width=1.9in]{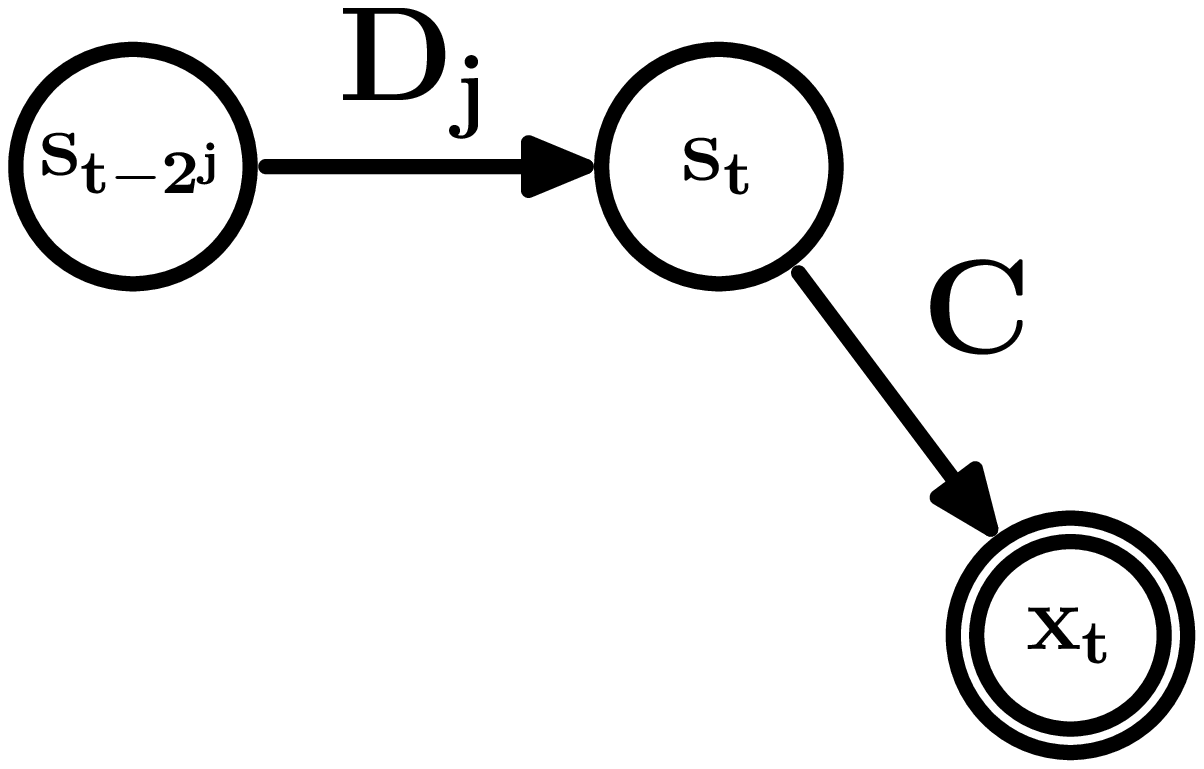}
\caption{\label{fig:def} {\bf Left Panel:} A state defining prediction. At training time, $\xx_{1}$
and $\xx_{2}$ are known. The essential goal is to predict $\xx_{2}$
given $\xx_{1}$ using bottleneck hidden variables $\ss_2$. Two distinct
mappings $A$ and $C$ are learned, with $\ss_2 \equiv A(\xx_{1})$.
{\bf Middle Panel:} A state evolution prediction. At training time,
$\xx_{t-1}$ and $\ss_{t-1}$ are used to predict $\ss_t$ via the 
operator $B(\xx_{t-1},\ss_{t-1})$ such that $\xx_t$
is reproduced via $C(\ss_t)$. 
{\bf Right Panel:} A state projection prediction. At training
time, $\ss_{t-2^{j}}$ is used to predict $\ss_{t}$ such that $\xx_t$
is reproduced via $C(\ss_{t})$ for $j\in\{0,1,2,....\left\lfloor \log_{2}T\right\rfloor \}$.
}
\vspace{-0.1in}
\end{figure*}

\section{Learning SPR-DM}

The essential idea of our algorithm is to use a bottlenecking approach
to construct an implicit definition of state, along with state space
evolution and projection operators to answer various natural questions
we might pose.

\subsection{Training}

There are two parts to understanding the training process. The first
is the architecture trained, and the second is the exact method of
training this architecture. Note that our architecture is essentially
functional rather than representational.

\subsubsection{Architecture}

Graphically, in order to recover the system dynamics, we solve two
distinct kinds of prediction problems. To understand these graphs it
is essential to understand that the arrows do \emph{not} represent
graphical models. Instead, they are a depiction of which information
is used to predict which other information. We distinguish
observations and hidden {}``state'' as double circles and circles respectively,
to make clear what is observed and what is not.

The first prediction problem solved in \Figref{fig:def}, left panel, provides our
initial definition of state. Essentially, state is {}``that
information which summarizes the first observation in predicting the
second observation''.  Compared to a conventional dynamic model, the
quantity $\ss_{2}$ may be a sufficient statistic of the state
posterior after integrating $x_1$, the posterior after integrating
$x_1$ and evolving one step or some intermediate mixture.  This
ambiguity is fundamental, but inessential.

The second prediction problem is state evolution, shown in \Figref{fig:def}, middle panel. 
Here, we use a state and an observation to
predict the next state, reusing the prediction of state from
observation from the first step.  Note that even though there are two
sources of information for predicting $\ss_{t}$, only one prediction
problem (using both sources) is solved. Operator $B$ is what is used
to integrate new information into the state of an online system.

%\begin{figure}
%\centering
% \includegraphics[width=1.9in]{Update2} 
%\caption{\label{fig:evolution} A state evolution prediction. At training time,
%$\xx_{t-1}$ and $\ss_{t-1}$ are used to predict $\ss_t$ via the operator $B(\xx_{t-1},\ss_{t-1})$ such that $\xx_t$
%is reproduced via $C(\ss_t)$. }
%\vspace{-0.1in} 
%\end{figure}

Without loss of generality, in the notation of 
\Figref{fig:def} %and \Figref{fig:evolution},
we consider $f_0(X_{t+1:\infty})=E[ \xx_{t+1} | X_{1:t}]$, and denote
$C^{f_0}$ by $C$. An alternative interpretation of $C$, which we do not 
distinguish in this paper, is to learn the probability distribution over $\xx_{t+1}$.
It should be understood that our algorithm can be applied
with other choices of $f_0$.

The above two learning diagrams are used to obtain the system dynamics
($A$ and $B$). One can then use the learned system dynamics to learn
prediction rules $C^f$ with any function $f$ of interest.  Here, we
consider the problem of predicting $\xx_{t+k}$ at different ranges of
$k=2^j$. This gives a state projection operator $D_j: \ss_t \to
\ss_{t+2^j}$, without observing the future sequence
$\xx_{t+1},\xx_{t+2},\cdots$.  The learning of state projection is
presented in \Figref{fig:def}, right panel.  The idea in state projection is
that we want to build a predictor of the observation far in the
future. To do this, we'll chain together several projection operators
from the current state. To make the system computationally more efficient,
we learn $\left\lfloor \log_{2}T\right\rfloor $ operators, each
specialized to cover different timespans.  Note that state evaluation
provides an efficient way to learn $\xx_{t+k}$ based on $\ss_t$
simultaneously for multiple $k$ through combination of projection
operators. If computation is not an issue, one may also learn
$\xx_{t+k}$ based on $\ss_t$ separately for each $k$.

%
%\begin{figure}
%\centering
% \includegraphics[width=1.9in]{Projection2} 
%\caption{\label{fig:projection} A state projection prediction. At training
%time, $\ss_{t-2^{j}}$ is used to predict $\ss_{t}$ such that $\xx_t$
%is reproduced via $C(\ss_{t})$ for $j\in\{0,1,2,....\left\lfloor \log_{2}T\right\rfloor \}$.}
%\vspace{-0.2in}
%\end{figure}

\subsubsection{Method}
Training of $A$ is straightforward. Training of $C$ is complicated
by the fact that samples appear at multiple timesteps, but otherwise
straightforward given the other components. To deal with multiple
timesteps, it is important for our correctness proof in section \ref{sub:correctness}
that the observation $\xx_{t}$ include the timestep $t$. The training
of $D$ is also straightforward given everything else (and again,
we'll require the timestep be a part of the update for the correctness
proofs).

The most difficult thing to train is $B$, since an alteration to $B$
can cascade over multiple timesteps. The method we chose takes
advantage of both local and global information to provide a fast
near-optimal solution.

\begin{enumerate}
\item {\bf Initialization}: Learn $B_{t},C_{t}$ starting from timestep
  $t=1$ and conditioning on the previous learned value. Multitask
  learning or initialization with prior solutions may be applied to
  improve convergence here.  In our experiments, we initialize
  $B_t,C_t$ to the average parameter values of previous timesteps and
  use stochastic gradient descent techniques for learning.
\item {\bf Conditional Training} Learn an alteration $B'$ which
  optimizes performance given that the existing $B_{t}$ are used at
  every other time step.  Since computational performance is an issue,
  we use a ``backprop through time'' gradient descent style algorithm.
  For each timestep $t$, we compute the change in squared loss for all
  future observations using the chain rule, and update according to
  the negative gradient.
\item {\bf Iteration}: Update $B$ using stochastic mixing according to
  $B_{i}=\alpha B'+(1-\alpha)B_{i-1}$ where $\alpha$ is the stochastic
  mixing parameter.  The precise method of stochastic mixing used in
  the experiments is equivalent to applying the derivative update with
  probability $\alpha$ and not update with probability $1-\alpha$, which is a
  computational and representational improvement over Searn~\cite{Searn}.
\end{enumerate}
We prove (below) that the method in step (1) alone is consistent.
Steps (2) and (3) are used to force convergence to a single $B$ and
$C$ while retaining the performance gained in step (1).  The intuition
behind step (3) is that when $\alpha=o(\frac{1}{T})$, with high
probability $B'$ is executed only once, implying that $B'$ need only
perform well with respect to the learning problem induced by the rest
of the system to improve the overall system. This approach was first
described in Conservative Policy Iteration~\cite{CPI}.

\subsection{Testing}

We imagine testing the algorithm by asking questions like: what is
the probability of observation $\xx_{t'}$ given what is known up to
time $t$ for $t'>t$? This is done by using $A(\xx_{1})$ to get $\ss_{2}$,
then using $B(\xx_i,\ss_{i})$ to evolve the state to $\ss_{t}$. Then
the time interval from $t'-t$ is broken down into factors of $2$,
and the corresponding state projection operators $D_{i}$ are applied
to the state resulting in a prediction for $\ss_{t'-1}$. This is transformed
into a prediction for $\xx_{t'}$ using operator $C$.

\section{Analysis}

\subsection{Computation}

The computational requirements depend on the exact training method
used.
For the initialization step, training of $A$, $B_{t}$, and $C_{t}$
requires just $O(nT)$ examples. Training $D_{i}$ can be done with just
$O(nT\log_{2}T)$ examples.  For the iterative methods, an extra factor
of $T$ is generally required per iteration for learning $B$.

\subsection{Consistency}

\label{sub:correctness}
We now show that under appropriate assumptions, the SPR-DM model can
be learned in the infinite sample limit using our algorithm. Due to
the space limitation, we only consider the non-agnostic situation,
where the SPR-DM model is exact.  That is, the functions $A$, $B$, $C$
used in our learning algorithm contains the correct functions. The
agnostic setting, where the SPR-DM model is only approximately correct,
can be analyzed using perturbation techniques (e.g., for linear
systems, this is done in \cite{Linear}).  Although such analysis is
useful, the fundamental insight is identical to the non-agnostic
analysis considered here.

We consider the following constraints in the SPR-DM model.  We assume
that the model is \emph{invertible}: The distribution over $\xx_t$
(more generally, the definition can be extended to other vector
functions $\phi_0(\xx_t,\ldots,)$) is a sufficient statistic for the
state $\ss_t$ that generates $\xx_t$.  This is a nontrivial limitation
of state based dynamic models which retains the ability to capture long
range dependencies.

\begin{defn}
(Invertible SPR-DM) The SPR-DM in Definition~\ref{def:spr-dm} is
  invertible if there exist a function $E$ such that for all
  $t$, $E( C^f( \ss_t ) ) = \ss_t$.
\end{defn}
Invertibility is a natural assumption, but it's important to understand
that invertible dynamic systems are a subset of dynamic systems 
as shown by the following hidden Markov model example:

%\begin{example}
{\bf Example 4.1} {\it A hidden Markov model which is not invertible}: 
%\end{example}
Suppose there are two
observations, $0$ and $1$ where the first observation is uniform
random, the second given the first is always $0$, and the third is
the same as the first. Under this setting, the two valid sequences
are $000$ and $101$. There is a hidden Markov model which is not
invertible that can express this sequence. In particular, suppose
state $s_{1}$ is $(0,1)$ or $(1,1)$ and state $s_{2}$ is $(0,2)$ or
$(1,2)$, with a conditional observation that is $P(0|*,1)=1$ and $P(0|0,2)=1$
and $P(0|1,2)=0$. However, no invertible hidden Markov model can
induce a distribution over these sequences because the distribution
on $x_{2}$ is always $0$, implying that a specification of state
is impossible due to lack of information. 

Although Invertible SPR-DMs form a limited subset of SPR-DMs,
they are still nontrivial as the following example
shows.

%\begin{example}
{\bf Example 4.2} 
{\it An Invertible hidden Markov model with long range dependencies}: Suppose
there are two observations $0$ and $1$ and two states $s_{1}$ and
$s_{2}$. Let the first observation always be $0$ and the first state
be uniform random $P(s_{1}|0)=P(s_{2}|0)=0.5$. Let the states only
self-transition according to $P(s_{1}|s_{1})=1$ and $P(s_{2}|s_{2})=1$.
Let the observations be according to the following distribution: $P(0|s_{1})=0.75$,
$P(0|s_{2})=0.25$. Given only one observation, the probability of
state $s_{1}$ is $0.75$ or $0.25$ for observations $0$ or $1$
respectively. Given $T$ observations, the probability of state $s_{1}$
converges to $0$ or $1$ exponentially fast in $T$ using Bayes Law
and the Chernoff bound.
%\end{example}

The above two examples illustrate the intuition behind invertibility.
One can extend the concept by incorporating look aheads: that is,
instead of taking $C$ as the probability of $\xx_{t}$ given $\ss_t$, we
may let $C$ be the probability of $X_{t:t+k}$ given
$\ss_t$. This broadens the class of invertible models.  In this
notation, invertibility means that if two states $\ss_t$ and $\ss_t'$
induce the same short range behavior $X_{t:t_k}$, then they
are identical in the sense they induce the same behavior for all
future observations: $X_{t+1:\infty}$.  Generally speaking,
non-invertible models are those that cannot be efficiently learned by
any algorithm because we do not have sufficient information to recover
states that have different long range dynamics but identical behavior
in short ranges. In fact, there are well-known hardness results for
learning such models in the theoretical analysis of hidden Markov
models.  There are no known efficient methods to capture non-trivial
long-range effects.  This implies that our restriction is not only
necessary, but also not a significant limitation in comparison to any
other known efficient learning algorithms.

Next we prove that our algorithm can recover any invertible hidden
Markov model given sufficiently powerful prediction with infinitely 
many samples. This is analogous to similar infinite-sample
consistency results for supervised learning. 

\begin{thm}
(Consistency) For all Invertible SPR-DMs, if all prediction problems
  are solved perfectly, then for all $i$, $p(\xx_{i}|\xx_{1},...,\xx_{i-1})$
  is given by:
  $\hat{C}(\hat{B}(\xx_{i-1},\hat{B}(\xx_{i-2},...,\hat{A}(\xx_{1})...)))$.
\end{thm}
A similar theorem statement holds for projections.
\vspace{-0.10in}
\begin{proof}
The proof is by induction.

The base case is $C(A(\xx_1)) = \hat{C}_{2}(\hat{A}(\xx_{1}))$ which
holds under the assumption that the prediction problem is solved
perfectly. In the inductive case, define: 
$s_2 = A(\xx_1)$, 
$\hat{s}_2 = \hat{A}(\xx_1)$, 
$\ss_i = B(\xx_{i-1},\ss_{i-1})$, 
$\hat{\ss}_i = \hat{B}_i(\xx_{i-1},\hat{\ss}_{i-1})$
and assume $C(\ss_i) = \hat{C}_i(\hat{\ss}_i)$.
Invertibility and the inductive assumption implies there exists $E$ such that:
$\ss_i = E(\hat{C}_i(\hat{\ss}_i))$. 
Consequently, 
there exists $\hat{C}_{i+1} = C $ and $\hat{B}_{i+1}(\xx_i,\hat{\ss}_i) = B(\xx_i,E(\hat{C}(\hat{\ss}_i)))$ such that:
\vspace{-0.08in}
\begin{align*}
C(B(\xx_i,\ss_i)) = \hat{C}_{i+1}(\hat{B}_{i+1}(\xx_i,\hat{\ss}_i)
\end{align*}
proving the inductive case.
\end{proof}

\section{Experiments}
In this section we present experimental results on two datasets 
that involve high-dimensional, highly-structured sequence data. 
The first dataset is the motion capture data 
that comes from CMU Graphics Lab Motion Capture Database.
The second dataset is the Weizmann dataset\footnote{
Available at 
http://www.wisdom.weizmann.ac.il/ \\ ~$\sim$vision/SpaceTimeActions.html.},
which contains video sequences of nine human subjects performing 
various actions.

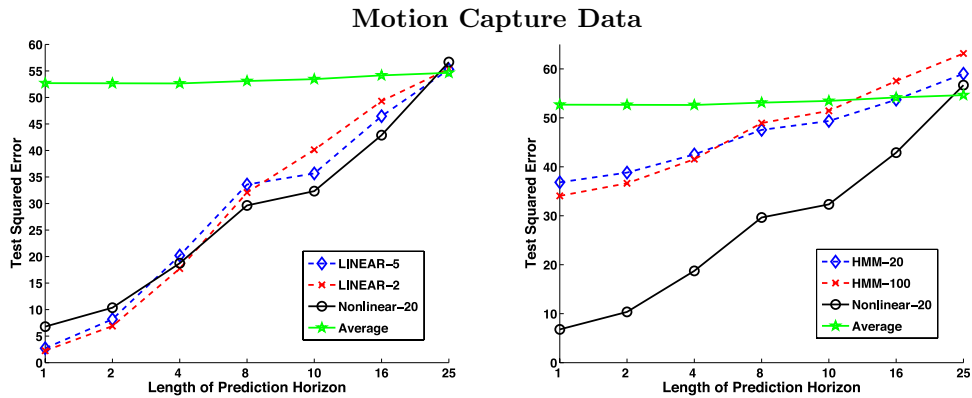
\begin{figure*}[t!]
\vspace{0.2in} 
\hbox{ \hspace{0.9in}
\includegraphics[width=2.4in]{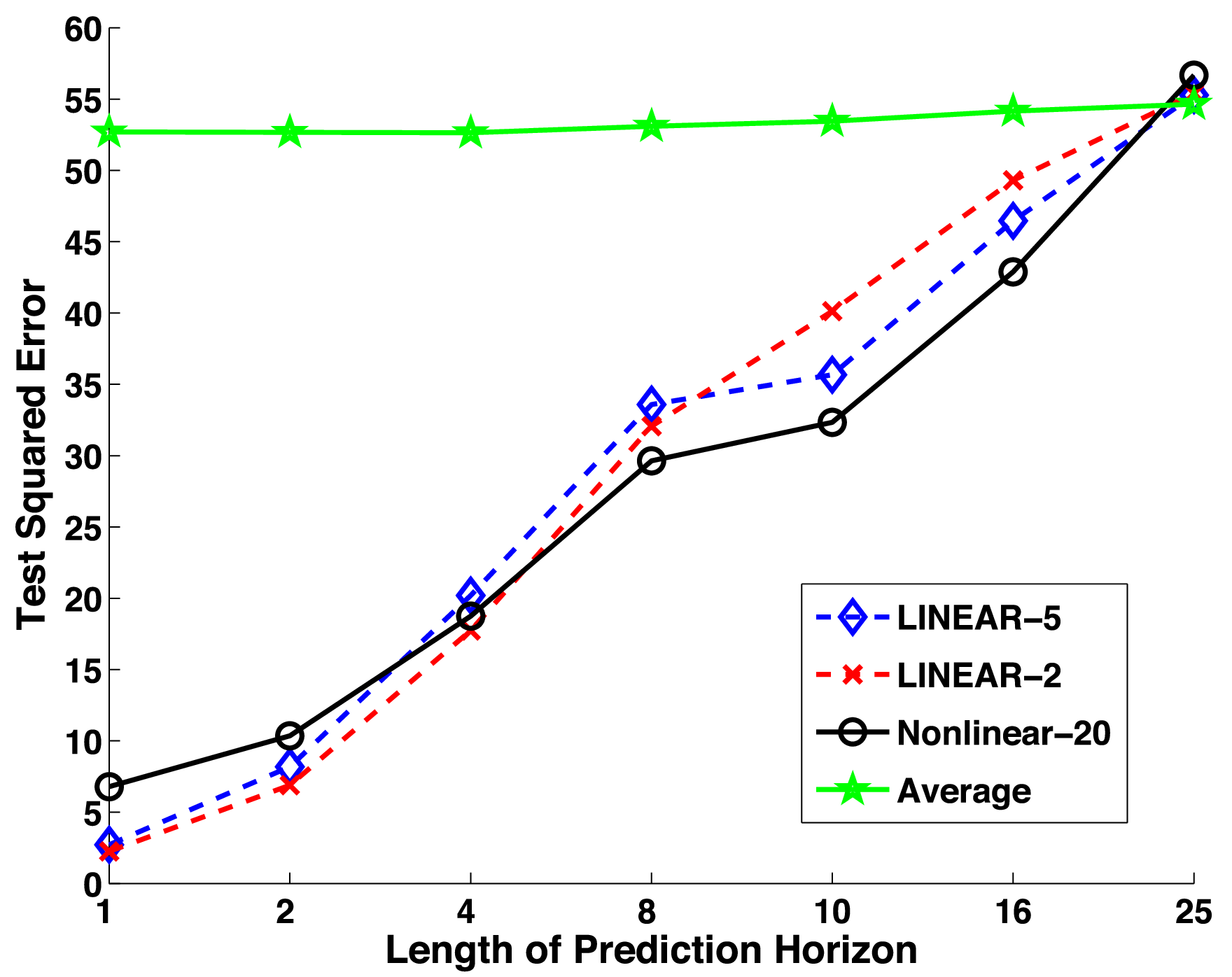}
\hspace{0.20in}
\includegraphics[width=2.4in]{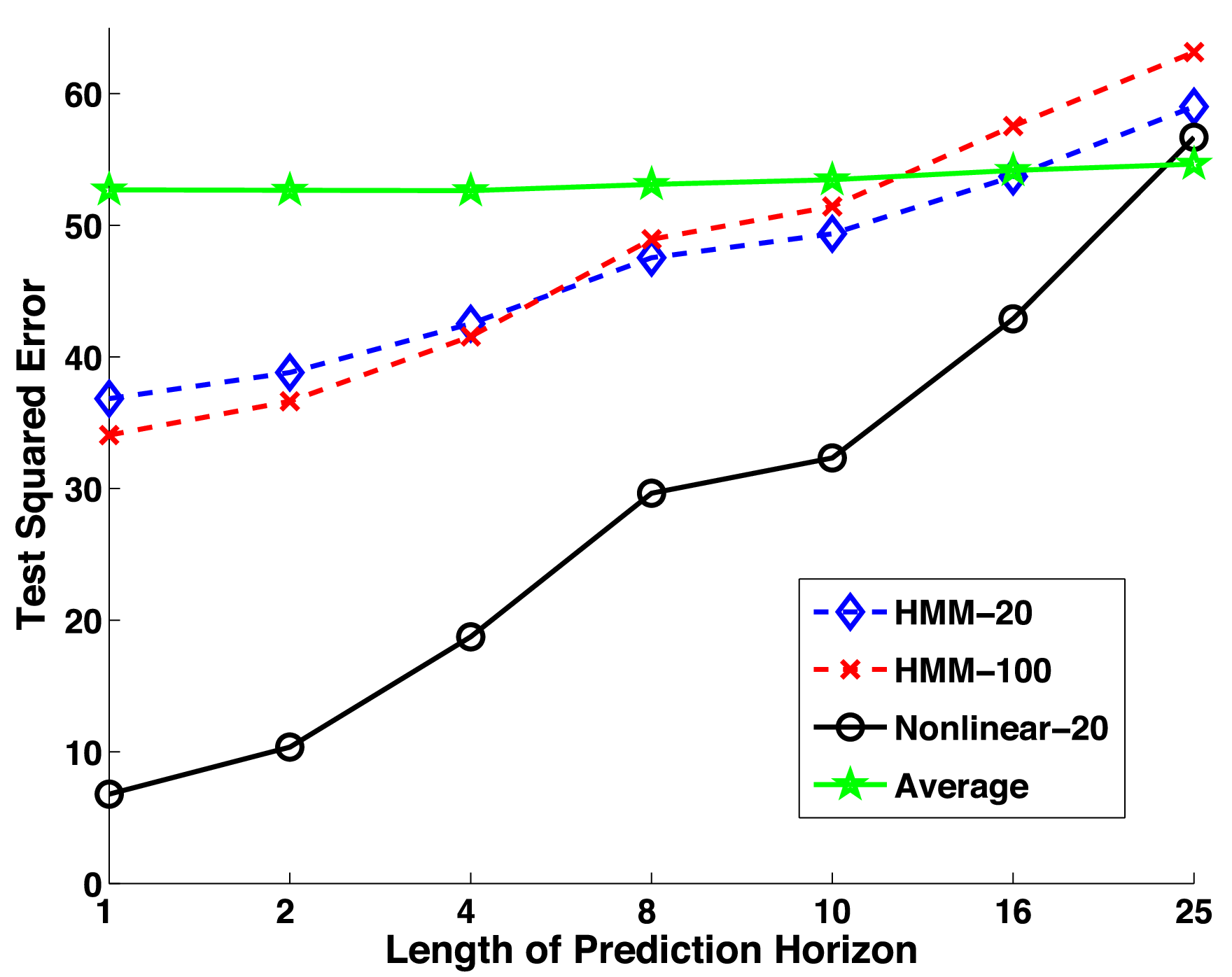}
\begin{picture}(0,0)(0,0)
\setlength{\epsfxsize}{0.5in}
\put(-240, +140){\textrm{{\bf Motion Capture Data}}}
\end{picture}
}
\caption{ \label{fig:mcap} \small
{\bf Left panel:} compares the average squared test error 
as a function of prediction horizon for three models:   
two linear autoregressive models when conditioning on 2 and 5 previous time
steps, and the nonlinear model that uses a 20-dimensional hidden state.
{\bf Right panel:} 
compares nonlinear model with 20-state and 100-state HMM models.  
The average predictor always predicts a vector of zeros. 
}
\vspace{-0.2in} 
\end{figure*}

\subsection{Details of Training}
While the introduced framework allows us to use many available
nonlinear supervised learning algorithms, in our experiments
we use the following parametric forms for our operators:
\beqa
  \ss_2 = &~ A(\xx_1) = \sigma\left(A^{\top}\xx_1 + \bb \right), \nonumber \\
  \ss_t = &~ B(\xx_{t-1},\ss_{t-1}) = \sigma\left(B^\top_1 \xx_{t-1} + B^\top_2 \ss_{t-1}+ \bb\right),  \nonumber \\
  \hat{\xx}_t = &~ C(\ss_t) = C^{\top} \ss_t + \aa, \nonumber \\
  \ss_{t+2^j} = &~ D_j(\ss_t) = D_j^{\top} \ss_t + \dd, \nonumber
\eeqa
where $\sigma(y) = 1/(1+\exp(-y))$ is the logistic function,
applied componentwise,
$\{C,B,A,Di_j,\aa,\bb,\dd\}$ are the model parameters with
$\aa,\bb$ and $\dd$ representing the bias terms.

For both datasets, during the initialization step,
the values of $\{B_t,C_t\}$  are initialized to the average parameter
values of previous timesteps\footnote{The values of $A,B_1,C_1$
were initialized with small random values sampled
from a zero-mean normal distribution with standard deviation of 0.01.
}.
Learning of $\{B_t,C_t\}$ then proceeds by minimizing
the squared loss using stochastic gradient descent.
For each time step, we use 500 parameter updates, with
learning rate of $0.001$.
We then used 500 iterations of
stochastic mixing, using gradients obtained by
backpropagation through time.
The stochastic mixing rate
$\alpha$ was set to 0.9 and
was gradually annealed towards zero.
We experimented with various values for the learning rate and
various annealing schedules for the mixing rate $\alpha$.
Our results are fairly robust to
variations in these parameters.
In all experiments we were conditioning on the
two previous time steps to predict the next.

\subsection{Motion Capture Data}
The human motion capture data consists of sequences of 3D joint angles
plus body orientation and translation.  The dataset was preprocessed
to be invariant to isometries \cite{conf/nips/Graham}, and contains
various walking styles, including normal, drunk, graceful, gangly,
sexy, dinosaur, chicken, and strong.  We split at random the data into
30 training and 8 test sequences, each of length 50.  The training
data was further split at random into the 25 training and 5 validation
sequences.  Each time step was represented by a vector of 58
real-valued numbers.  The dataset was also normalized to have zero
mean was scaled by a single number, so that the variance across each
dimension was on average equal to 1.  The dimensionality of the hidden
state was set to 20.

\begin{figure*}[t!]
\vspace{0.2in} 
\hbox{ \hspace{0.9in}
\includegraphics[width=2.4in]{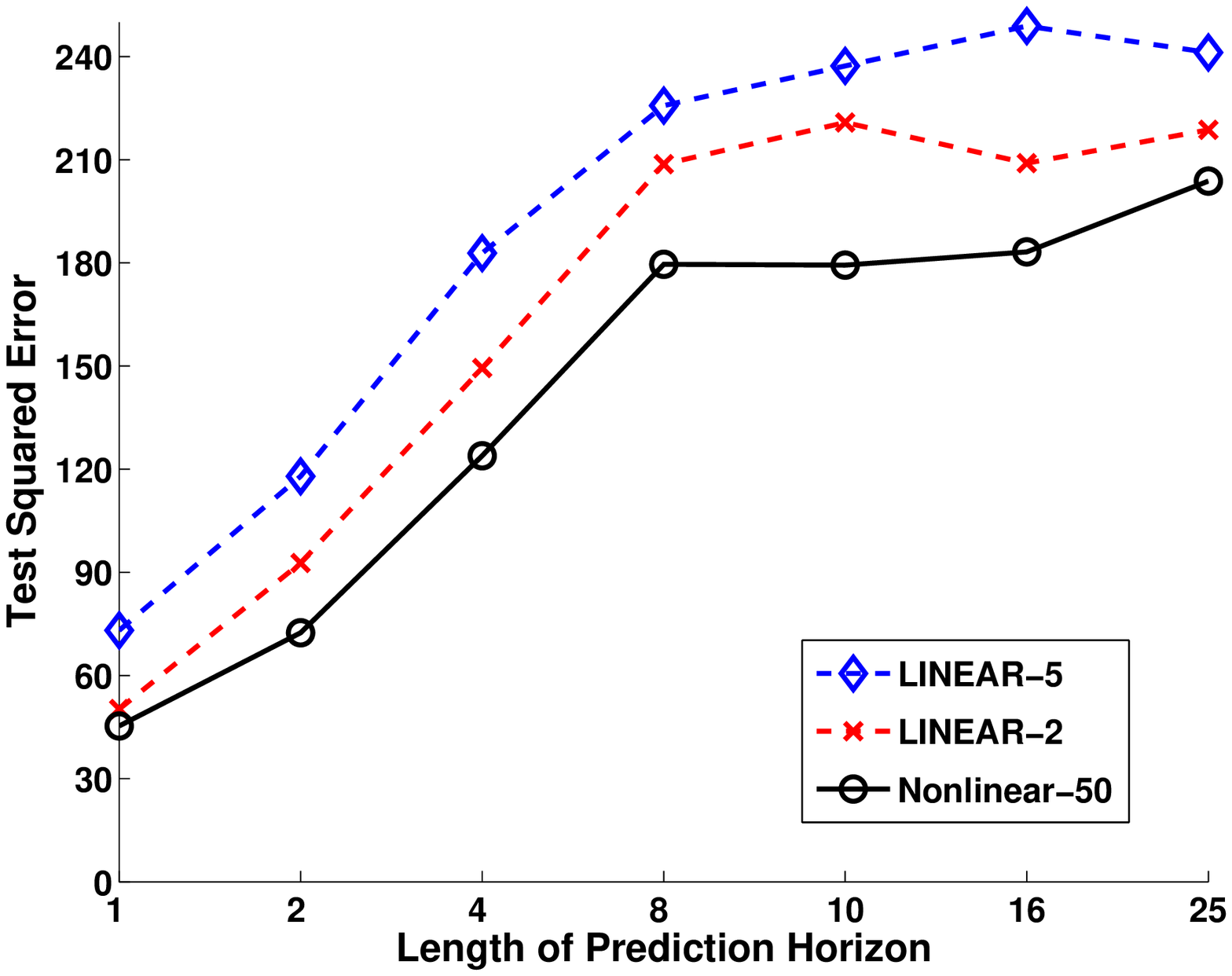}
\hspace{0.20in}
\includegraphics[width=2.4in]{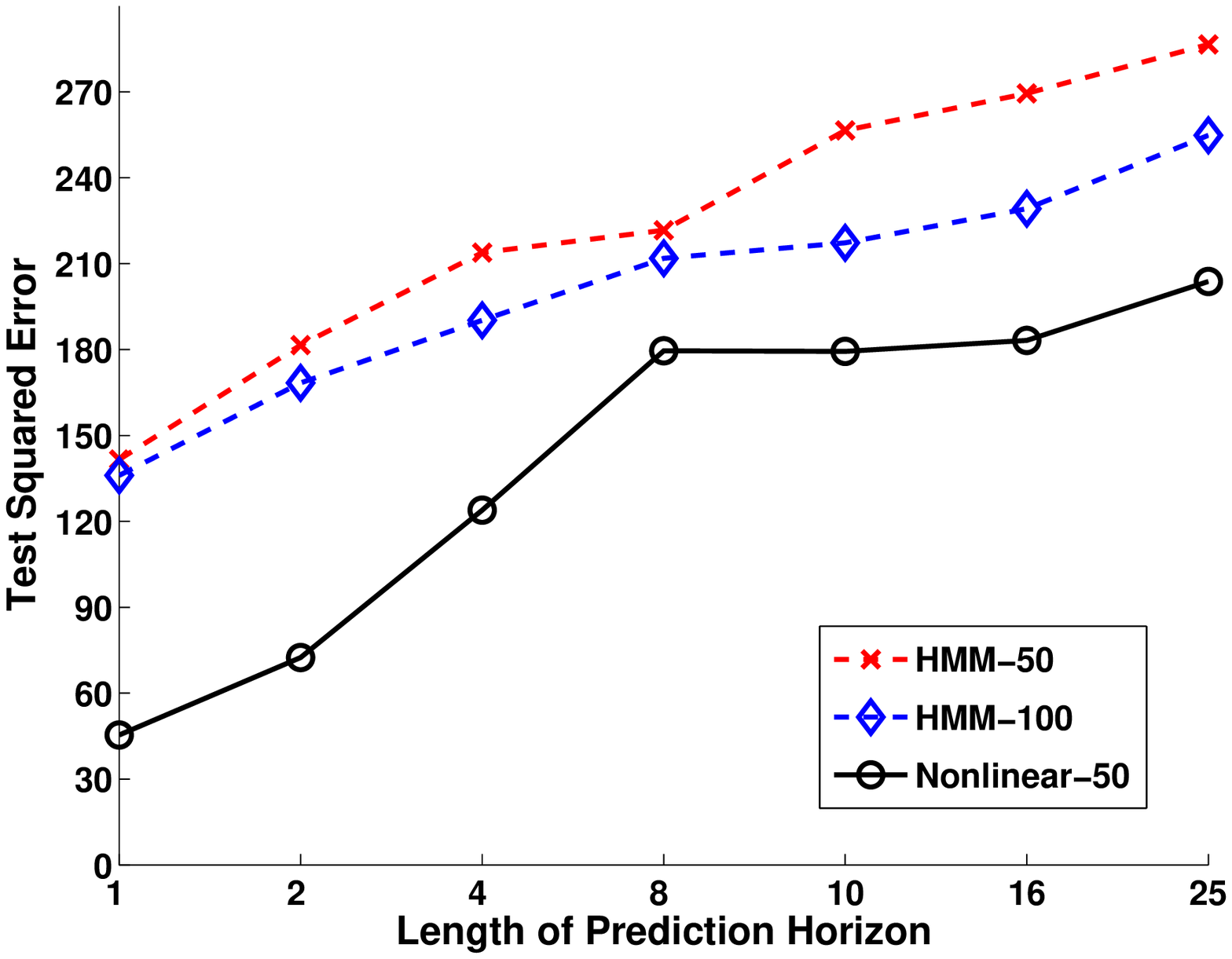}
\begin{picture}(0,0)(0,0)
\setlength{\epsfxsize}{0.5in}
\put(-220, +140){\textrm{{\bf Video Data}}}
\end{picture}
}
\caption{ \label{fig:video} \small
{\bf Left panel:} compares the average squared test error
for three models:
two linear autoregressive models, and the nonlinear
model that uses a 50-dimensional hidden state.
{\bf Right panel:}
compares nonlinear model to 50-state and 100-state HMM models.
}
\vspace{-0.2in} 
\end{figure*}

Figure~\ref{fig:mcap} shows the average test prediction errors 
using squared loss,
where the prediction horizon ranges over 1,2,4,8,10,16, and 25.
The nonlinear model was compared to two simple autoregressive linear models
that operate directly in the input space.
The first linear model, LINEAR-2, makes predictions $\hat{x}_{t+k}$ via the linear 
combination of the two previous time steps:
\beqa
 \hat{\xx}_{t+k} = {L^1}^\top\xx_t + {L^2}^\top\xx_{t-1} + \ll. 
\eeqa 
The model parameters $\{L^1,L^2,\ll\}$ were fit by 
ridge regression. The second model, LINEAR-5, makes
predictions by conditioning on the previous five time steps.    
We note that the number of
the model parameters for these simple autoregressive linear models 
grows linearly with the input information.
Hence when faced with high-dimensional sequence data,
learning linear operators directly in the input space 
is unlikely to perform well.

It is interesting to observe that autoregressive linear models
perform quite well in terms of making short-range predictions.
This is probably due to the fact that locally, motion capture data 
is linear. However, the nonlinear model performs considerably better 
compared to both linear models when making long-range predictions.  
Figure \ref{fig:mcap} (right panel) further shows that the proposed nonlinear model
performs considerably better than 20 and 100-state HMM's. 
Both HMM's use Gaussian distribution as their observation model.  
It is obvious that a simple HMM model is unable to 
cope with complex nonlinear dynamics. Even a 100-state 
HMM is unable to generalize. 

\subsection{Modeling Video}
Results on the motion capture dataset show that a nonlinear model can
outperform linear and HMM models, when making long-range predictions.
In this section we present results on the Weizmann dataset, which is
considerably more difficult than the motion capture dataset.

The Weizmann dataset contains video sequences of nine human subjects
performing various actions, including waving one hand, waving two hands,
jumping, and bending.  Each video sequence was preprocessed by placing
a bounding box around a person performing an action. The dataset was
then downsampled to $29 \times 16$ images, hence each time step was
represented by a vector of 464 real-valued numbers.  We split at random the data
into into 36 training (30 training and 6 validation), and 10 test
sequences, each of length 50.  The dataset was also normalized to have
zero mean and variance 1. The dimension of the hidden state was set to
50.

Figure \ref{fig:video} shows that the nonlinear model consistently
outperforms both linear autoregressive and HMM models, particularly when
making long-range predictions.  It is interesting to observe that on
this dataset, the nonlinear model outperforms the autoregressive model
even when making short-range predictions.  

\section{Conclusions} 
In this paper we introduced a new approach to learning nonlinear
dynamical systems and showed that it performs well on rather hard
high-dimensional time series datasets compared to standard models such
as HMMs or linear predictors.  We believe that the presented framework
opens up an entirely new set of devices for nonlinear dynamic
modeling.  It removes several obstacles in the traditional approach
that requires heavy human design, and allows well-established
supervised learning algorithms to be used automatically for nonlinear
dynamic models.
\vspace{-0.1in} 

\section*{Acknowledgments} 
This work was done when Ruslan Salakhutdinov visited Yahoo.  Tong
Zhang is partially supported by NSF grant DMS-0706805.

\bibliography{nonline}
\vspace{-0.1in} 
\bibliographystyle{mlapa}

\end{document}